\documentclass[runningheads]{llncs}
\usepackage{graphicx}
\usepackage{bm}
\usepackage{amsmath}
\usepackage{amssymb}
\usepackage{multirow}
\usepackage{makecell}
\usepackage{marvosym}
\begin{document}
%
\title{SRR-Net: A Super-Resolution-Involved Reconstruction Method for High Resolution MR Imaging}
\titlerunning{SRR-Net: An SR-involved Reconstruction Method for HR MRI}
%

\author{Wenqi Huang\inst{1,3} \and
Sen Jia\inst{2} \and
Ziwen Ke\inst{1,3} \and
Zhuo-Xu Cui\inst{1} \and
Jing Cheng\inst{2} \and
Yanjie Zhu\inst{2} \and
Dong Liang\inst{1,2} (\Letter) 
}

\authorrunning{W. Huang et al.}
%
\institute{Research Center for Medical AI, Shenzhen Institute of Advanced Technology,
Chinese Academy of Sciences, Shenzhen, Guangdong, China\\
\email{\{wq.huang1, dong.liang\}@siat.ac.cn}\\
\and
Paul C. Lauterbur Research Center for Biomedical Imaging, Shenzhen
Institute of Advanced Technology, Chinese Academy of Sciences, Shenzhen,
Guangdong, China\\
\and
Shenzhen College of Advanced Technology, University of Chinese Academy
of Sciences, Shenzhen, Guangdong, China\\
}


%
\maketitle              
\begin{abstract}
Improving the image resolution and acquisition speed of magnetic resonance imaging (MRI) is a challenging problem. There are mainly two strategies dealing with the speed-resolution trade-off: (1) $k$-space undersampling with high-resolution acquisition, and (2) a pipeline of lower resolution image reconstruction and image super-resolution. However, these approaches either have limited performance at certain high acceleration factor or suffer from the error accumulation of two-step structure. In this paper, we combine the idea of MR reconstruction and image super-resolution, and work on recovering HR images from low-resolution under-sampled $k$-space data directly. Particularly, the SR-involved reconstruction can be formulated as a variational problem, and a learnable network unrolled from its solution algorithm is proposed. A discriminator was introduced to enhance the detail refining performance. Experiment results using in-vivo HR multi-coil brain data indicate that the proposed SRR-Net is capable of recovering high-resolution brain images with both good visual quality and perceptual quality.

\end{abstract}
\section{Introduction}
High-resolution (HR) MR imaging enables accurate depiction of the anatomical information in various clinical applications. However, if not under sampled, the considerable amount of image voxel will lead to unacceptable long scan time than regular MRI. Recent studies dealing with these problems can be generally divided into two categories: the high-resolution acquisition approach and the super-resolution (SR) approach.

The high-resolution acquisition approach treats HR MR imaging no difference from regular MR Imaging. This approach utilizes parallel imaging and compressed sensing techniques to reconstruct HR images from highly under-sampled $k$-space data. Early studies explored the priors in MR image for fast imaging, such as sparsity and low-rankness \cite{2006_cs,2010_pics,2015_NLD,2015_L_plus_S}. However, it is challenging to determine the regularization parameters of CS-methods. The acceleration factor is also very limited for acceptable image quality. More recently, deep learning has demonstrated tremendous success in improving imaging quality and speed of MRI \cite{2017_DCCNN,2018_CRNN,2019_DIMENSION}.
Nevertheless, these reconstruction methods are not intendedly designed for high-resolution MRI. The acceleration factor (AF) for a near-perfect reconstruction is still too low for high-resolution imaging due to a large amount of $k$-space points to be sampled. Moreover, higher resolution typically reduces SNR.

\begin{figure}[t]
    \centering
    \includegraphics[width=\textwidth]{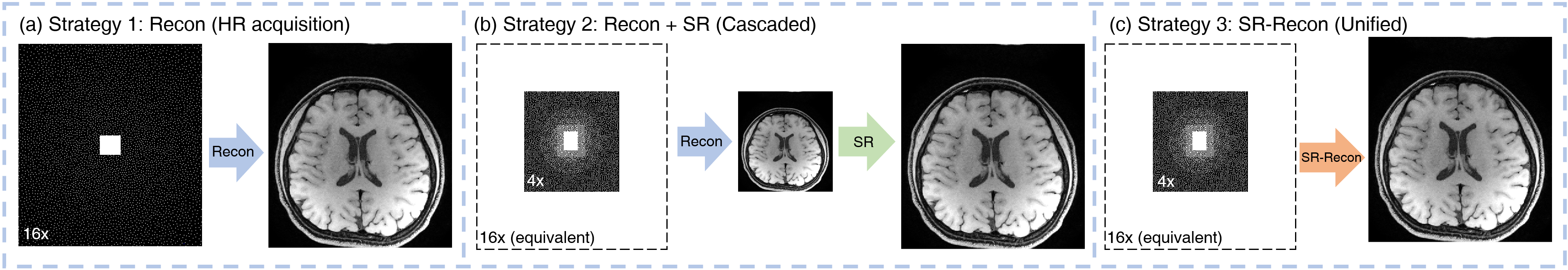}
    \caption{(a)(b) shows the existing acceleration strategies for high-resolution MRI, and (c) is the proposed strategy of SRR-Net. Strategy 1: direct reconstruction from under-sampled high-resolution acquisition data; Strategy 2: two-step reconstruction+super-resolution; Strategy 3: direct super-resolution-involved reconstruction. For clarity, we show the sampling mask instead of under sampled $k$-space data.}
    \label{fig:strategy}
\end{figure}

Image super-resolution is an alternative. First, a low-resolution MR image is reconstructed from $k$-space data sampled with a small sampling matrix. Then high-resolution images are generated using super-resolution methods. At the same equivalent acceleration factor (sampled points /target image voxels), this approach samples more densely in the low-frequency part of the $k$-space, which makes the high-quality reconstruction of the low-resolution image possible. Previous works of this approach c focuses on the super-resolution part. Like many super-resolution studies in computer vision field, these MRI super-resolution works usually takes LR images that are down sampled or blurred from HR images in image domain as input\cite{2017_ISBI_BrainSR,2018_ISBI_BrainSR_Dense,2018_MICCAI_MRISR,2018_DeepResolve,2020_TCI_MRISR}. The image domain degrading, which produces a completely different k-space data, is not practical in real-world MRI scanning. In addition, because there is no interaction between the low-resolution image reconstruction and the super-resolution, the information loss in the reconstruction step will cause error accumulation in the second step.

To make the best of both approaches and eliminate their drawbacks, we propose to realize MRI super-resolution and reconstruction in one step. For this purpose, we put forward a generative adversarial network termed as SRR-Net. The diagrams of all the three strategies are shown in Fig.~\ref{fig:strategy}. The SRR-Net is a unrolled network that reconstruct high-resolution MR images from low-resolution $k$-space data directly. The proposed SRR-Net avoided the information loss of two-step super-resolution. It is natural to combine the reconstruction process with the super-resolution process together because the common super-resolution networks, which usually using residual/dense blocks, share a similar structure to unrolling reconstruction methods. We compared the proposed SRR-Net with existing high-resolution acquisition methods and super-resolution methods at a 16-fold equivalent acceleration factor. Experiment results indicate that images reconstructed directly from the under-sampled low-resolution $k$-space data with SRR-Net have better visual and quantitative results. This work provides a new insight into high-speed, high-resolution MR imaging.

\section{Methodology}
\subsection{The iterative form of super-resolution-involved reconstruction}
The super-resolution MR imaging model can be formed as:
\begin{equation}
    y=\mathcal{MHFC}x + b
    \label{Eq:SRmodel}
\end{equation}
where $\mathcal{M}$ is the under-sampling operator, $\mathcal{F}$ is Fourier transform and $\mathcal{C}$ is the sensitivity maps. $x \in \mathbb{C}^N$ is the high-resolution image, and $b$ is the noise term. $y\in \mathbb{C}^M$ is the under-sampled low-resolution $k$-space data. The difference between Eq.~\ref{Eq:SRmodel} and the traditional MRI imaging model is the $k$-space down-sampling operator $\mathcal{H}:\mathbb{C}^{N}\rightarrow\mathbb{C}^{M}$, which cuts off the high-frequency component of the high-resolution image's $k$-space. This operation involves the adjustment of the resolution parameter $\Delta x$ in commercial MR scanners, which is more practical than image-domain down sampling.
For simplicity, we note $\mathcal{MHFC}$ as $A$. For solving inverse problem Eq.~\ref{Eq:SRmodel}, regularization is necessary, i.e, 
\begin{equation}
    \min_{x} R(x), \quad s.t.\quad y=Ax + b
    \label{Eq:inverse_problem}
\end{equation}
where $R$ denotes the regularizer. Numerically, Eq.~\ref{Eq:inverse_problem} can be solved by various optimization algorithms. We choose one that can give a simple and clear solution form. By introducing an auxiliary variable $s=x$, the penalty function of Eq.~\ref{Eq:inverse_problem} reads: 
we can get the penalty function of Eq.~\ref{Eq:inverse_problem}
\begin{equation}
    J(x,s) = \frac{1}{2}\|Ax-y\|_2^2+R(s)+\frac{\rho}{2}\|s-x\|_2^2
    \label{Eq:penalty}
\end{equation}
where $\frac{\rho}{2}$ is a penalty parameter.
Problem Eq.~\ref{Eq:penalty} can be solved via the following alternating minimization steps
\begin{equation}
    \left\{\begin{aligned}
    s_{k+1} &= \arg\min_{s}J(x_k,s)\\
    x_{k+1} &= \arg\min_{x}J(x,s_{k+1})
    \end{aligned}\right.
    \label{Eq:min_J}
\end{equation}
Since the calculation of $(A^*A+\rho I)^{-1}$ is very time consuming, we attempt to solve the $x$-subproblem inexactly. Specifically, for the $x$-subproblem at Eq.~\ref{Eq:min_J}, we can approximate the cost
function $J$ via linearizing the data fidelity $F(x) :=\frac{1}{2}\|Ax-y\|^2_2$ at $s_{k+1}$, i.e.,
\begin{equation}
    x_{k+1} = \arg\min_{x}\tilde{J}_{k+1}(x,s_{k+1})
\end{equation}
where
\begin{equation}
    \tilde{J}_{k+1}(x,s_{k+1}):=<\nabla F(s_{k+1}), x-s_{k+1}>+\frac{1}{2\eta}\|x-s_{k+1}\|_2^2+R(s_{k+1})
\end{equation}
Thus the subproblems in Eq.~\ref{Eq:min_J} can be formulated as
\begin{equation}
    \left\{\begin{aligned}
    s_{k+1} &= \arg\min_{s}\frac{\rho}{2}\|s-x_{k}\|_2^2+R(s)\\
    x_{k+1} &= \arg\min_{x}\frac{1}{2\eta}\|x-s_{k+1}\|_2^2+<\nabla F(s_{k+1}), x>
    \end{aligned}\right.
\end{equation}
Each of the subproblems can be solved via proximal gradient method and get the following update steps:
\begin{equation}
    \left\{\begin{aligned}
    s_{k+1} &= Prox_{R}(x_k)\\
    x_{k+1} &= s_{k+1} - \eta \nabla F(s_{k+1})
    \end{aligned}\right.
    \label{Eq:iterative_solution}
\end{equation}
where $Prox_{R}$ is a proximal operator depending on the regularizer $R$; $\eta$ is an update step size. Because we use a symbol $R$ to represent a generalized regularizer, the closed form of $Prox_{R}$ can not be formulated.
The $x$-update step is a data consistency operation, where $\nabla F(s)$ is the image-correction term of $k$-space residual. The gradient term is
\begin{equation}
    \nabla F(s)=\mathcal{C^*F^*H^*M^*(MHFC}(s)-y)
\end{equation}
in which $\mathcal{C^*, F^*, H^*}$ and $\mathcal{M^*}$ represents the adjoint operators of $\mathcal{C,F,H}$ and $\mathcal{M}$. Specifically, $\mathcal{H}^*\in \mathbb{C}^{M}\rightarrow\mathbb{C}^{N}$ is the up-sampling matrix, which applies zero paddings to the low resolution $k$-space to obtain the high-resolution image size.

\subsection{Super-Resolution-involved Reconstruction Network}
Like many other unrolling algorithms, the iteration steps in Eq.~\ref{Eq:iterative_solution} are unrolled into several iteration blocks and formulated the super-resolution-involved reconstruction network. Each of the blocks contains two network modules, which are named as refinement layer $S^k$ and data consistency layer $X^k$, corresponding to the $s$- and $x$-update steps in Eq.~\ref{Eq:iterative_solution}:

\begin{equation}
    \left\{\begin{aligned}
    S^{k+1}: s_{k+1}&= C(x_k)+\alpha\cdot x_k\\
    X^{k+1}: x_{k+1}&= s_{k+1} - \gamma \nabla F(s_{k+1})
    \end{aligned}\right.
\end{equation}

The refinement layer $S^k$ in Eq.(13) is designed as a scaled residual 3D convolution neural network to learn a customized proximal operator. $C$ is a 3-layer convolution neural network and $\alpha$ controls the residual scale. Both the residual scale factor $\alpha$ and step size $\gamma$ are learnable parameters (initialized with 1.0), bringing more flexibility to the model.

\begin{figure}
    \centering
    \includegraphics[width=0.9\textwidth]{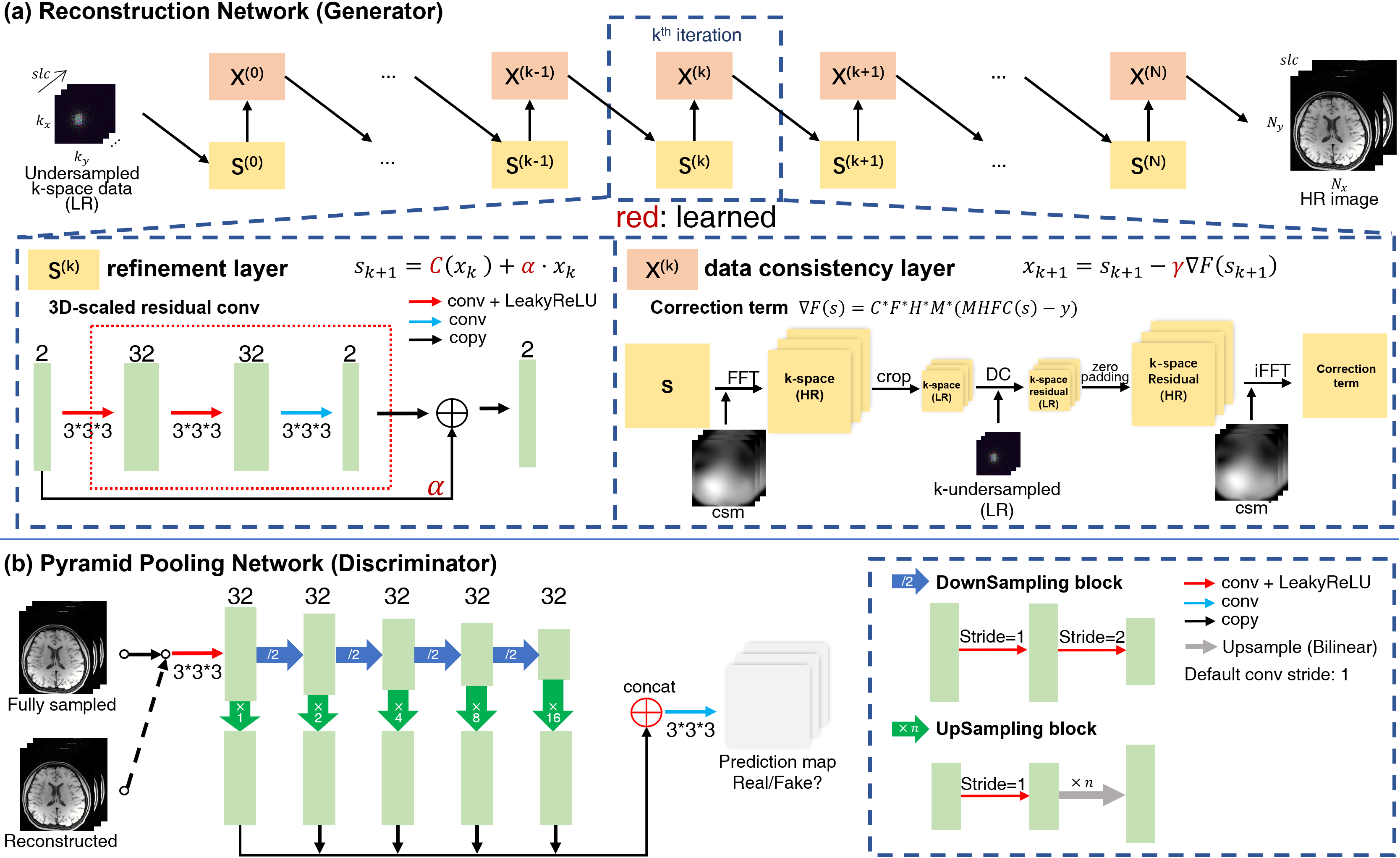}
    \caption{The schematic architecture of the proposed SRR-Net, which includes (a) a reconstruction network (Generator) and (b) a pyramid pooling network (Discriminator). HR: High-Resolution, LR: Low-Resolution, csm: coil sensitivity maps.}
    \label{fig:network}
\end{figure}
The most straightforward training objective for a reconstruction or super-resolution model is the voxel-wise difference between model output and the ground-truth image. While the $L_1$ or $L_2$ loss only focuses on the minimization of the average intensity error, rather than the sharpness and validity of restored structures, which will produce blurred details in the output image. 
To achieve better visual result of the reconstruction images, we introduce the idea of adversarial learning \cite{2014_GAN} to this super-resolution-involved reconstruction model. We use a pyramid pooling network \cite{2020_pyramid_pooling_disciminator} as the discriminator because its good capability to simultaneously capture details on different scales. This pair of generator and discriminator is trained by minimizing a modified WGAN-GP loss\cite{2017_WGAN_GP}
\begin{equation}
    \begin{aligned}
    L_D&=\mathbb{E}[D(\tilde{x})]-\mathbb{E}[D(x)] + \lambda \mathbb{E}[\|\nabla_{\hat{x}}D(\hat{x})\|_2-1)^2],\\
    L_G&=-\mathbb{E}[D(\tilde{x})] + \eta\mathbb{E}[\|x-\tilde{x}\|_2^2]
    \end{aligned}
\end{equation}
where $D$ and $G$ denote discriminator and generator, respectively. $\tilde{x}$ and $x$ is a pair of reconstructed image and label image. $\hat{x}$ is an interpolation of $x$ and $\tilde{x}$.
During the training stage, the discriminator compete with the reconstruction network (generator), driving the latter to recover more realistic visual details. After training, the reconstruction network takes the under-sampled low-resolution $k$-space data as input and outputs high-resolution images. Fig.~\ref{fig:network} shows the structure of the SRR-Net. 


\section{Experiments and Results}
\paragraph{\textbf{Data Preparation:}}
Fully sampled high-resolution whole-brain imaging was performed on five healthy volunteers with IRB-approved and informed consent. The data was acquired by the T1 weighted three-dimensional MATRIX sequence on a 3T scanner (uMR790, United Imaging Healthcare, China) with a commercial 32-channel head coil. Common imaging parameters include: sagittal imaging orientation, resolution=isotropic 0.6 mm$^3$, image size$=368\times 336\times 280$, TR/TE = 850/13 ms
. The fully-sampled $k$-space data was lossless compressed to 18 channels for efficient memory usage. A 1D Fourier transform was then performed along the readout dimension to decompose the single 3D data into multiple 2D slices. 
For Strategy 1, the data was retrospectively under sampled using a 16x Poisson disk sampling mask. For Strategy 2 and 3, the $k$-space data was center-cropped to $168\times 140$, and under sampled with a 4x Poisson disk mask to simulate a low-resolution acquisition. Because the center-cropped $k$-space is 4x smaller than before and the target is still $336\times 280$, the equivalent acceleration factor is still 16-fold. The $24\times 24$ $k$-space center of each slice was used to estimate the sensitivity maps with ESPIRiT. The complex-valued label images were combined using such sensitivity maps. We used the data from 3 subjects (1104 slices) for training and the rest (736 slices) for testing. Training data augmentation is conducted by striding (stride=4) along the slice dimension to create $8\times 336\times 280$ small 3D blocks. 

\paragraph{\textbf{Model and Parameters:}}
The main structure of the proposed SRR-Net is shown in Fig.~\ref{fig:network}. In the experiments, the reconstruction network is implemented with 10 iteration blocks, each of them has independent learnable parameters and convolution layers. The convolution blocks $C$ in the refinement blocks is a residual CNN with 3 convolution layers. We concatenate the real and imaginary part of each input of $C$ in the channel dimension for complex value convolution. The first two layers of the convolution block has 32 kernels, and the last one have 2 kernels to output the real and imaginary part, respectively. The kernel size is $3\times 3\times 3$. Leaky Rectifier Linear Units (Leaky ReLU) \cite{2013_leakyReLU} were selected as the nonlinear activation functions following the first two convolution layers. The $S^0$ is initialized with the zero filling image. The initial values for the learned residual factor $\alpha$ and the learned step size $\gamma$ in Eq.(17) are both set to 1.0. 

We trained the model with a batch size of 1 due to GPU memory limitation. The optimizer is ADAM and the exponential decay learning rate \cite{2012_exponential_decay_lr} was used in the training procedure, and the initial learning rate for both generator and discriminator was set to 0.001 with a decay of 0.95. The training procedure stopped after 50 epochs. The models were implemented with TensorFlow, and the training and testing is done on a GPU server with an Nvidia RTX 8000 Graphics Processing Unit (GPU, 48GB memory). It took approximately 30 hours for 50 epochs to train the network. The code will be released on GitHub once the paper is published. For a fair comparison, we tried our best to adjust all the methods mentioned in this paper to their best performance. 

\paragraph{\textbf{Evaluation:}}
\begin{table}[tbp]
    \centering
    \caption{Experiment explanations. Recon: Reconstruction, SR: Super-Resolution, Equav.AF: Equavelent Acceleration Factor, HR/LR: using high/low-resolution under-sampled $k$-space data as input. \#Param: the amount of model parameters.}
    \begin{tabular}{c|c|c|c|c|c}
        \hline \hline
        Abbreviation & Strategy & Recon Method & SR Method & Equiv. AF & \#Param\\
        \hline
        PICS & 1 & \multicolumn{2}{c|}{$l_1$-ESPIRiT} & 16(HR) & - \\
        \hline
        ISTA-Net & 1 & \multicolumn{2}{c|}{ISTA-Net} & 16(HR) & 330K\\
        \hline
        PICS-KI & 2 & $l_1$-ESPIRiT  & $k$-space interp. & 16(LR) & -\\
        \hline
        ISTA-Net-KI & 2 & ISTA-Net & $k$-space interp. & 16(LR) & 330K\\
        \hline
        PICS-EDSR & 2 & $l_1$-ESPIRiT & EDSR & 16(LR) & 40.7M\\
        \hline
        ISTA-Net-EDSR & 2 & ISTA-Net & EDSR & 16(LR) & 330K+40.7M\\
        \hline
        SRR-Net-NoADV & 3 & \multicolumn{2}{c|}{SRR-Net without discriminator} & 16(LR) & 311K\\
        \hline
        SRR-Net & 3 & \multicolumn{2}{c|}{SRR-Net with discriminator} & 16(LR) & 311K+361K\\
        \hline \hline
    \end{tabular}
    \label{tab:exp_explaination}
\end{table}

\begin{figure}[htb]
    \centering
    \includegraphics[width=0.8\textwidth]{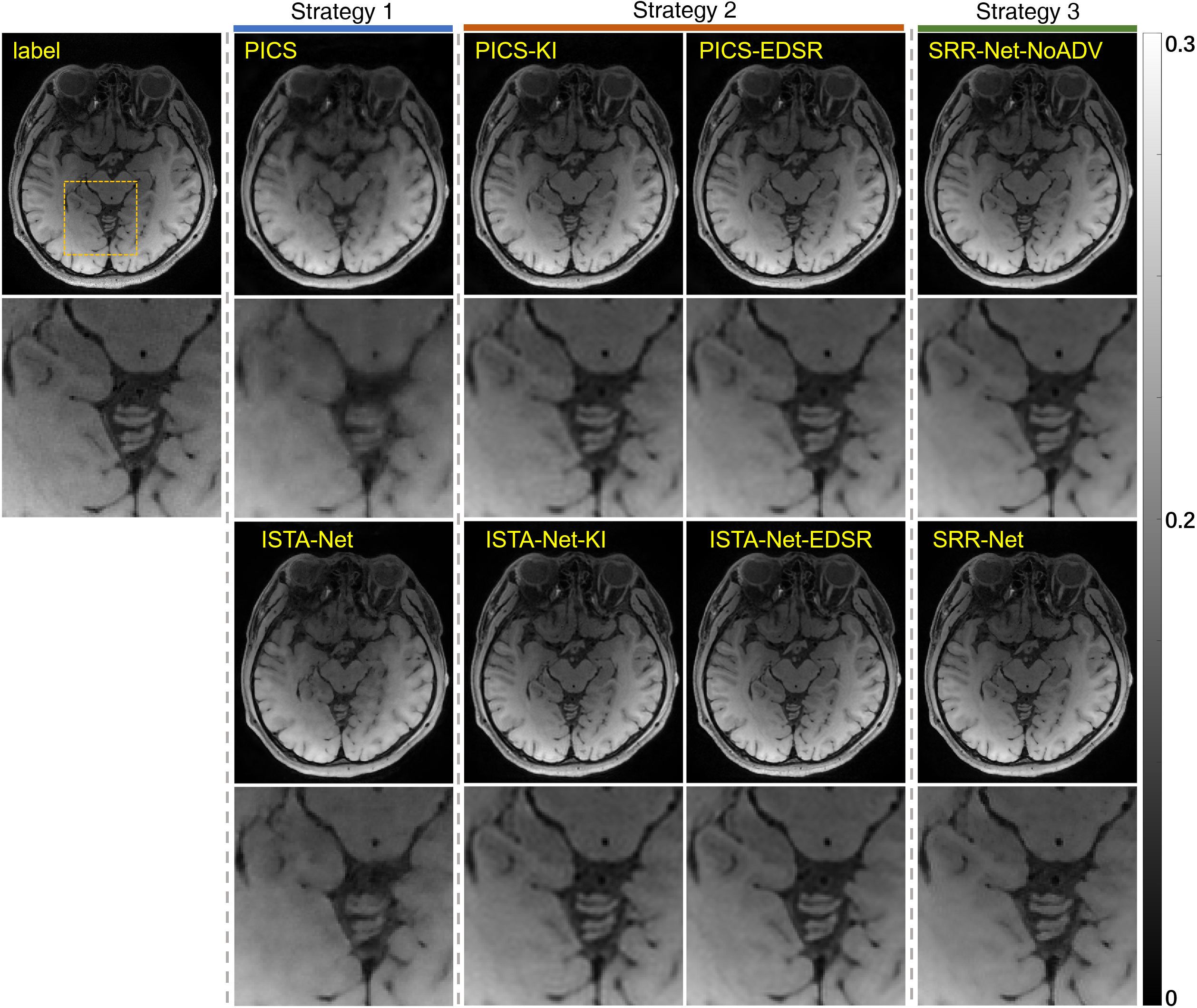}
    \caption{The visual results of different methods. The zoom-in ROIs of the enclosed part are provided in the second row.}
    \label{fig:SRresults}
\end{figure}

\begin{table}[tb]
    \centering
    \caption{Quantitative results of all experiments.}
    \begin{tabular}{c|c||c|c|c|c|c|c|c|c}
        \hline \hline
        Subj. & Metric & PICS & \makecell[c]{ISTA\\-Net} & \makecell[c]{PICS\\-KI} & \makecell[c]{ISTA-Net\\-KI} & \makecell[c]{PICS\\-EDSR} & \makecell[c]{ISTA-Net\\-EDSR} & \makecell[c]{SRR-Net\\-NoADV} & SRR-Net \\

        \hline
        \multirow{5}{*}{1} & PSNR & 40.16 & 41.20 & 42.67 & 43.02 & 42.76 & \textbf{43.46} & 43.25 & 42.39 \\
        \cline{2-10}
        & SSIM & 0.9331 & 0.9561 & 0.9563 & 0.9711 & 0.9565 & \textbf{0.9762} & 0.9720 & 0.9696\\
        \cline{2-10}
        & NIQE & 3.29 & 3.81 & 4.08 & 4.35 & 3.71 & 4.13 & 3.96 & \textbf{3.23}\\ 
        \cline{2-10}
        & PI & 6.05 & 6.51 & 6.79 & 7.05 & 6.43 & 6.83 & 6.67 & \textbf{5.92}\\ 
        
        \hline
        \multirow{5}{*}{2} & PSNR & 41.65 & 42.72 & 44.14 & 44.40 & 44.28 & 44.06 & \textbf{44.69} & 44.01\\
        \cline{2-10}
        & SSIM & 0.9499 & 0.9680 & 0.9703 & 0.9784 & 0.9708 & \textbf{0.9802} & 0.9794 & 0.9777 \\
        \cline{2-10}
        & NIQE & 3.59 & 3.72 & 4.18 & 4.32 & 3.93 & 3.86 & 4.16 & \textbf{3.26}\\ 
        \cline{2-10}
        & PI & 6.37 & 6.42 & 6.88 & 7.03 & 6.65 & 6.56 & 6.86 & \textbf{5.95}\\ 
        \hline
        \multicolumn{10}{l}{Better: PSNR: higher; SSIM: higher, NIQE: lower; PI: lower.}
    \end{tabular}
    \label{tab:metrics}
\end{table}
To quantitatively measure SRR-Net's performance, we use PSNR and SSIM as reference-based metrics. We also evaluate the perceptual quality via Naturalness Image Quality Evaluator (NIQE) \cite{2012_NIQE} and Perceptual Index (PI) \cite{2018_SR_challenge}, which are commonly used in image super-resolution challenges. For clarity, all the experiments conducted in this paper are listed in Tabel.~\ref{tab:exp_explaination}.



Fig.~\ref{fig:SRresults} shows the visual comparison of different methods. For Strategy 1, we select two reconstruction methods: $l_1$-ESPIRiT \cite{2014_ESPIRiT}, a traditional parallel imaging compressed sensing (PICS) algorithm, and ISTA-Net\cite{2018_ISTANET}, a deep learning reconstruction algorithm. We can observe that at 16-fold acceleration, both the reconstruction methods failed with unacceptable blur and artifacts. It demonstrates that direct reconstruction of highly under-sampled high-resolution data is very challenging for both CS methods and DL methods. 

For Strategy 2, we choose $l_1$-ESPIRiT and ISTA-Net for LR image reconstruction, and $k$-space interpolation \cite{2001_KI} and EDSR \cite{2017_EDSR} for super-resolution. The third and forth columns of Fig.~\ref{fig:SRresults} show the 4 combinations of these methods. The LR $k$-space is sampled with 4-fold acceleration, which is equivalent to a 16-fold acceleration for HR images.
It is worth notion that $k$-space interpolation is a widely used super-resolution method in clinical applications. It performs better than traditional image-domain interpolation algorithms such as Bi-linear or Cubic interpolation. The reconstruction quality of PICS-KI and ISTA-Net-KI is close, which means the LR images of the two methods are similar. While PICS-EDSR shows a more blurry result than ISTA-Net-EDSR because of the error accumulation of the two-step strategy. 
Both ISTA-Net-EDSR and the proposed SRR-Net are able to recover fine details, and the SRR-Net is slightly sharper. In addition, the number of parameters of SRR-Net, shown in Table~\ref{tab:exp_explaination}, is much smaller than EDSR methods. The improvement not only comes from the elimination of error accumulation, but also from the adversarial scheme. We eliminated the discriminator in SRR-Net, and formulated an SRR-Net-NoADV for comparison. 
The adversarial scheme enables sharper edges and better feature restoration.
The quantitative results listed in Table.~\ref{tab:metrics} also corresponds to the visual impression. Although SRR-Net does not have the best PSNR and SSIM results, its perceptual metrics and visual quality beat all other methods. It indicates that, at extremely high acceleration, super-resolution-involved reconstruction is a good alternative to high-resolution acquisition methods and two-stage SR methods.


\section{Conclusion}
This paper developed and evaluated a super-resolution-involved reconstruction network, termed as SRR-Net, for 3D high-resolution brain MR imaging. We demonstrated that the proposed SRR-Net outperforms existing reconstruction methods and two-step recon-SR methods at the same equivalent acceleration factor, which shows its potential in fast high-resolution MR imaging. We also presented that with GAN-guided training, SRR-Net has better detail recovery and higher perceptual quality. This work serves as a preliminary attempt to bridge the gap between MR image reconstruction and super-resolution. The idea of super-resolution-involved reconstruction allows a 16-fold reduction of scan time with promising image details and perceptual quality, which would improve the image quality and scan speed of high-resolution MR imaging. More validations and techniques of GAN-guided SR-involved reconstruction can be investigated in future works.
\\

%
%
%
\bibliographystyle{unsrt}
\bibliography{ref}
%




\end{document}